\documentclass[sigconf]{acmart}
\usepackage{graphicx}
\usepackage{amsmath}
\usepackage{amsthm}

\usepackage{amssymb}
\usepackage{amsfonts}
\usepackage{bm}
\usepackage{caption}
\usepackage{multirow}
\usepackage{subfigure}
\usepackage{enumitem}
\usepackage{array}
\usepackage{algorithm}
\usepackage{algorithmic}
\theoremstyle{definition}

\usepackage{colortbl}

\AtBeginDocument{%
  \providecommand\BibTeX{{%
    \normalfont B\kern-0.5em{\scshape i\kern-0.25em b}\kern-0.8em\TeX}}}

\copyrightyear{2022} 
\acmYear{2022} 
\setcopyright{acmlicensed}\acmConference[CIKM '22]{Proceedings of the 31st ACM International Conference on Information and Knowledge Management}{October 17--21, 2022}{Atlanta, GA, USA}
\acmBooktitle{Proceedings of the 31st ACM International Conference on Information and Knowledge Management (CIKM '22), October 17--21, 2022, Atlanta, GA, USA}
\acmPrice{15.00}
\acmDOI{10.1145/3511808.3557294}
\acmISBN{978-1-4503-9236-5/22/10}

\begin{document}

\title{Domain Adversarial Spatial-Temporal Network: \\A Transferable Framework for Short-term Traffic Forecasting across Cities}

\author{Yihong Tang}
\authornote{These authors contributed equally to this work.}
\affiliation{%
  \department{Department of Urban Planning and Design}
  \institution{University of Hong Kong}
  \city{Hong Kong SAR}
  \country{China}
}
 \email{yihongt@connect.hku.hk}
 
\author{Ao Qu}
 \authornotemark[1]
\affiliation{%
  \department{Laboratory for Information and Decision Systems}
  \institution{Massachusetts Institute of Technology}
  \city{Boston}
  \country{USA}
}
 \email{qua@mit.edu}
 
\author{Andy H.F. Chow}
\affiliation{
  \department{Department of Advanced Design and Systems Engineering}
  \institution{City University of Hong Kong}
  \city{Hong Kong SAR}
  \country{China}
}
\email{andychow@cityu.edu.hk}

\author{William H.K. Lam}
\affiliation{
  \department{Department of Civil and Environmental Engineering}
  \institution{The Hong Kong Polytechnic University}
  \city{Hong Kong SAR}
  \country{China}
}
\email{william.lam@polyu.edu.hk}

\author{S.C. Wong}
\affiliation{
  \department{Department of Civil Engineering}
  \institution{University of Hong Kong}
  \city{Hong Kong SAR}
  \country{China}
}
\email{hhecwsc@hku.hk}

\author{Wei Ma}
 \authornotemark[1]
\authornote{Corresponding author.}
\affiliation{%
  \department{Department of Civil and Environmental Engineering}
  \department{Research Institute for Sustainable Urban Development}
  \institution{The Hong Kong Polytechnic University}
  \city{Hong Kong SAR}
  \country{China}
}

\email{wei.w.ma@polyu.edu.hk}

\renewcommand{\shortauthors}{Yihong, et al.}

\begin{abstract}
Accurate real-time traffic forecast is critical for intelligent transportation systems (ITS) and it serves as the cornerstone of various smart mobility applications. 
Though this research area is dominated by deep learning, recent studies indicate that the accuracy improvement by developing new model structures is becoming marginal. Instead, we envision that the improvement can be achieved by transferring the ``forecasting-related knowledge'' across cities with different data distributions and network topologies.
To this end, this paper aims to propose a novel transferable traffic forecasting framework:  Domain Adversarial Spatial-Temporal Network (\textsc{DastNet}). \textsc{DastNet} is pre-trained on multiple source networks and fine-tuned with the target network's traffic data. Specifically, we leverage the graph representation learning and adversarial domain adaptation techniques to learn the domain-invariant node embeddings, which are further incorporated to model the temporal traffic data.
To the best of our knowledge, we are the first to employ adversarial multi-domain adaptation for network-wide traffic forecasting problems.
\textsc{DastNet} consistently outperforms all state-of-the-art baseline methods  on three benchmark datasets. 
The trained \textsc{DastNet} is applied to Hong Kong's new traffic detectors, and accurate traffic predictions can be delivered immediately (within one day) when the detector is available.
Overall, this study suggests an alternative to enhance the traffic forecasting methods and provides practical implications for cities lacking historical traffic data. Source codes of \textsc{DastNet} are available at 
\url{https://github.com/YihongT/DASTNet}.
\end{abstract}




\maketitle

\section{Introduction}
\label{sec:intro}

\begin{figure}[t]
    \centering
    \includegraphics[width=.39\textwidth]{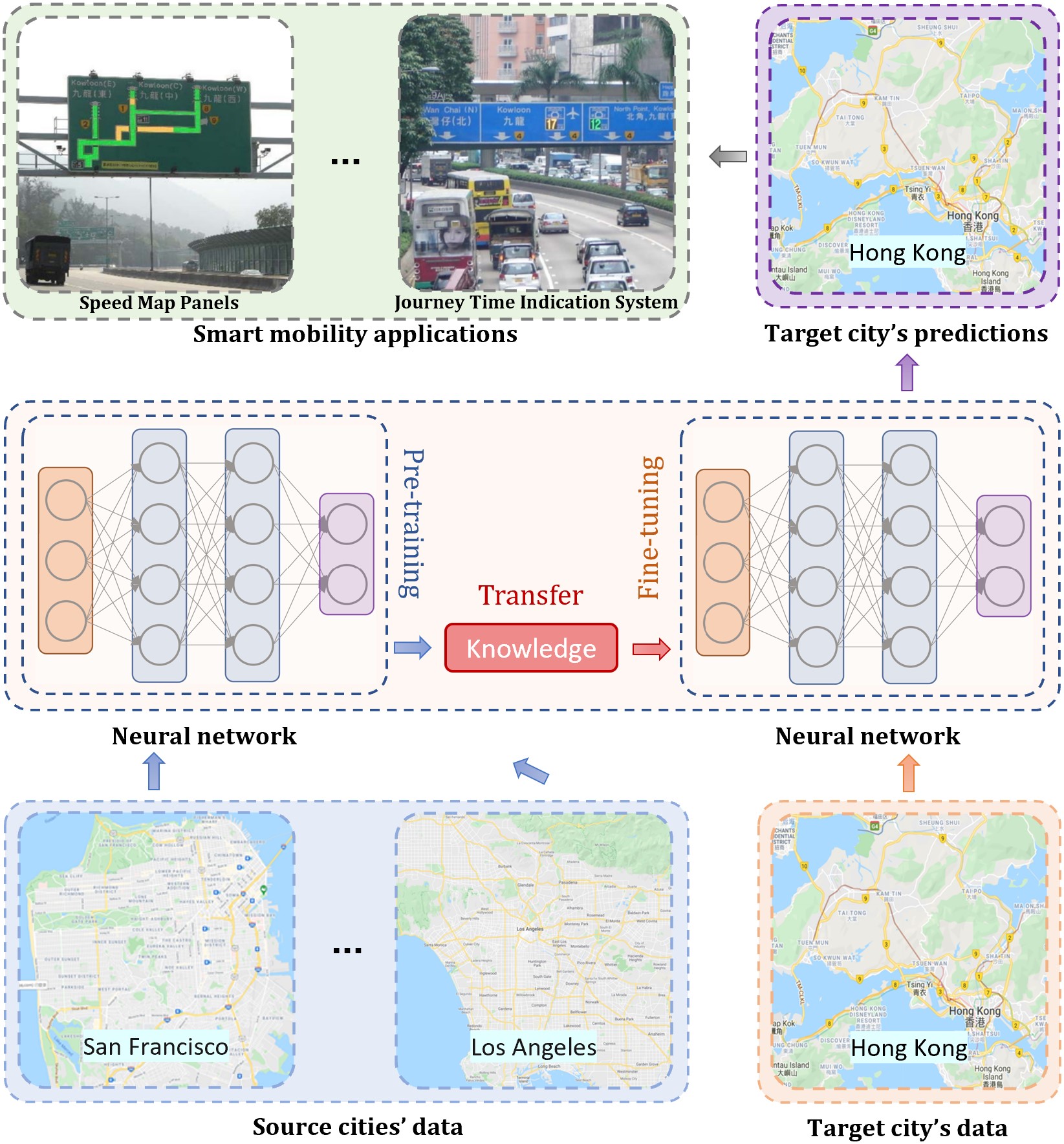}
    \caption{An overview of the transferable traffic forecasting problem and its applications.} 
    \label{fig:illustration}
\end{figure}

Short-term traffic forecasting \cite{jiang2021graph,bolshinsky2012traffic} has always been a challenging task due to the complex and dynamic spatial-temporal dependencies of the network-wide traffic states. When the spatial attributes and temporal patterns of traffic states are convoluted, their intrinsic interactions could make the traffic forecasting problem intractable. 
Many classical methods \cite{williams2003modeling,drucker1997support} take temporal information into consideration and cannot effectively utilize spatial information. 
With the rise of deep learning and its application in intelligent transportation systems (ITS) \cite{dai2020hybrid,barnes2020bustr,zhang2020curb}, a number of deep learning components, such as convolutional neural networks (\textsc{Cnn}s) \cite{o2015introduction}, graph neural networks (\textsc{Gnn}s) \cite{kipf2016semi}, and recurrent neural networks (\textsc{Rnn}s) \cite{fu2016using}, are employed to model the spatial-temporal characteristics of the traffic data \cite{song2020spatial,du2017traffic,guo2019attention,cai2020traffic,li2021dynamic}. These deep learning based spatial-temporal models achieve impressive performances on traffic forecasting tasks.


However, recent studies indicate that the improvement of the forecasting accuracy induced by modifying neural network structures has become marginal \cite{jiang2021graph}, and hence it is in great need to seek alternative approaches to further boost up the performance of the deep learning-based traffic forecasting models. One key observation for current traffic forecasting models is that: most existing models are designed for a single city or network. Therefore, a natural idea is to train and apply the traffic forecasting models across multiple cities, with the hope that the ``knowledge related to traffic forecasting'' can be transferred among cities, as illustrated in Figure~\ref{fig:illustration}. The idea of transfer learning has achieved huge success in the area of computer vision, language processing, and so on \cite{pan2009survey,patel2015visual,liu2019survey}, while the related studies for traffic forecasting are premature \cite{yin2021deep}.

There are few traffic forecasting methods aiming to adopt transfer learning to improve model performances across cities \cite{wang2018cross,pan2020spatio,yao2019learning,wang2021spatio,tian2021transfer}. These methods partition a city into a grid map based on the longitude and latitude, and then rely on the transferability of \textsc{Cnn} filters for the grids. However, the city-partitioning approaches overlook the topological relationship of the road network while modeling the actual traffic states on road networks has more practical value and significance. 
The complexity and variety of road networks' topological structures could result in untransferable models for most deep learning-based forecasting models \cite{mallick2021transfer}. Specifically, we consider the road networks as graphs, and the challenge is to effectively map different road network structures to the same embedding space and reduce the discrepancies among the distribution of node embedding with representation learning on graphs.

\begin{figure}[t]
    \centering
    \includegraphics[width=.4\textwidth]{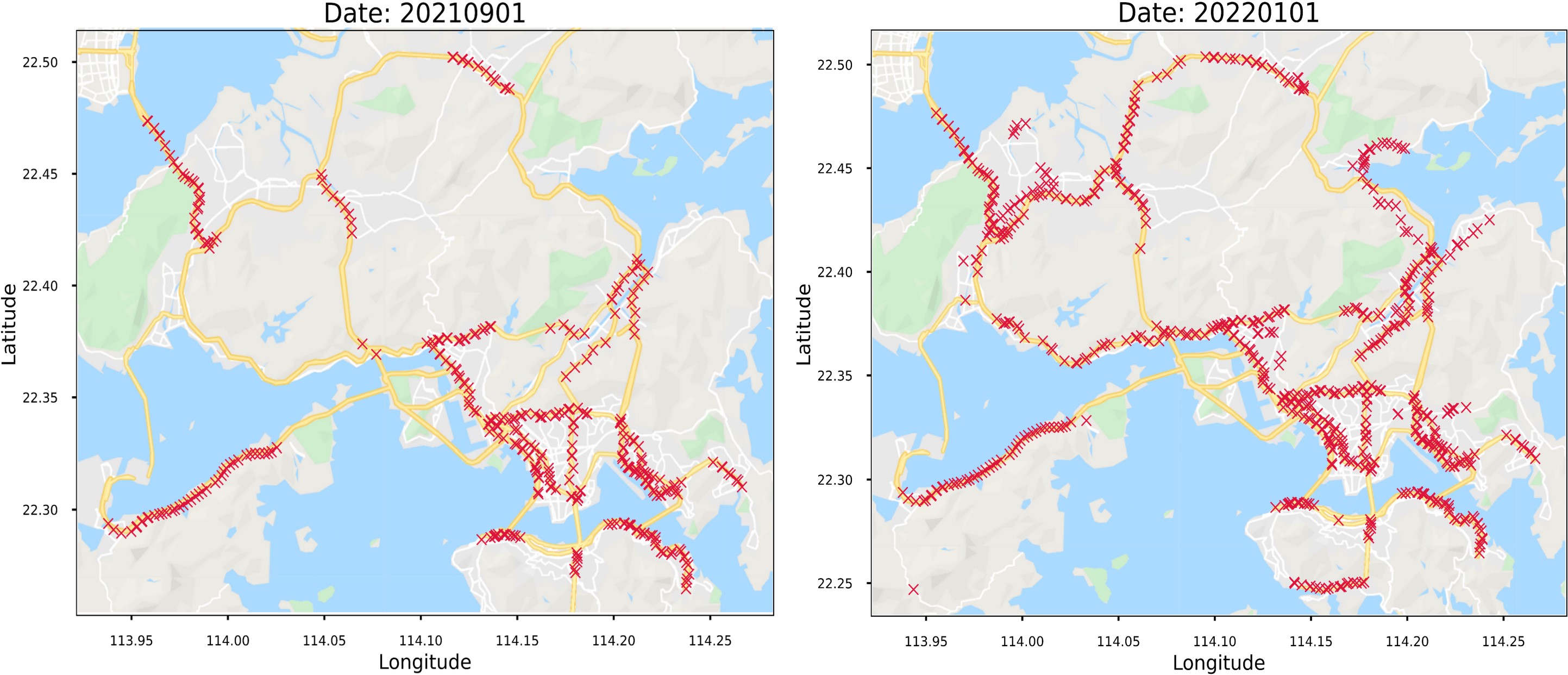}
    \caption{Available detectors in Hong Kong in September 2021 (left) and January 2022 (right).}
    \label{fig:HKsensor}
\end{figure}

As a practical example, Hong Kong is determined to transform into a smart city. The Smart City Blueprint for Hong Kong 2.0 was released in December 2020, which outlines the future smart city applications in Hong Kong \cite{HKsmart}. Building an open-sourced \textit{traffic data analytic platform} is one essential smart mobility application among those applications. Consequently, Hong Kong's Transport Department is gradually releasing the traffic data starting from the middle of 2021 \cite{HKdata}. As the number of detectors is still increasing now (as shown in Figure \ref{fig:HKsensor}), the duration of historical traffic data from the new detectors can be less than one month, making it impractical to train an existing traffic forecasting model. This situation also happens in many other cities such as Paris, Shenzhen, and Liverpool \cite{vivacitylabs2022}, as the concept of smart cities just steps into the deployment phase globally. One can see that a successful transferable traffic forecasting framework could enable the smooth transition and early deployment of smart mobility applications.

To summarize, it is both theoretically and practically essential to develop a network-wide deep transferable framework for traffic forecasting across cities. 
In view of this, we propose a novel framework called Domain Adversarial Spatial-Temporal Network (\textsc{DastNet}), which is designed for the transferable traffic forecasting problem. 
This framework maps the raw node features to node embeddings through a spatial encoder. The embedding is induced to be domain-invariant by a domain classifier and is fused with traffic data in the temporal forecaster for traffic forecasting across cities. Overall, the main contributions of our work are as follows:
\begin{itemize}[leftmargin=*]
    \item We rigorously formulate a novel transferable traffic forecasting problem for general road networks across cities.
    \item We develop the domain adversarial spatial-temporal network (\textsc{DastNet}), a transferable spatial-temporal traffic forecasting framework based on multi-domains adversarial adaptation. 
    To the best of our knowledge, this is the first time that the adversarial domain adaption is used in traffic forecasting to effectively learn the transferable knowledge in multiple cities.
    \item We conduct extensive experiments on three real-world datasets, and the experimental results show that our framework consistently outperforms state-of-the-art models.
    \item The trained \textsc{DastNet} is applied to Hong Kong's newly collected traffic flow data, and the results are encouraging and could provide implications for the actual deployment of Hong Kong's traffic surveillance and control systems such as Speed Map Panels (SMP) and Journey Time Indication System (JTIS) \cite{tam2011application}. 
\end{itemize}
The remainder of this paper is organized as follows. Section \ref{sec:relatedworks} reviews the related work on spatial-temporal traffic forecasting and transfer learning with deep domain adaptation. Section \ref{sec:preliminaries} formulates the transferable traffic forecasting problem. Section \ref{sec:method} introduces details of \textsc{DastNet}. In section \ref{sec:exp}, we evaluate the performance of the proposed framework on three real-world datasets as well as the new traffic data in Hong Kong. We conclude the paper in Section \ref{sec:conclusion}.

\section{Related Works}\label{sec:relatedworks}

\subsection{Spatial-Temporal Traffic Forecasting}

The spatial-temporal traffic forecasting problem is an important research topic in spatial-temporal data mining and has been widely studied in recent years. 
Recently, researchers utilized \textsc{Gnn}s \cite{kipf2016semi,velivckovic2017graph,zhang2020spatial,xie2020deep,park2020st} to model the  spatial-temporal networked data since \textsc{Gnn}s are powerful for extracting spatial features from road networks. Most existing works use \textsc{Gnn}s and \textsc{Rnn}s to learn spatial and temporal features, respectively \cite{zhao2019t}. \textsc{Stgcn} \cite{yu2017spatio} uses \textsc{Cnn} to model temporal dependencies. \textsc{Astgcn} \cite{guo2019attention} utilizes attention mechanism to capture the dynamics of spatial-temporal dependencies. \textsc{Dcrnn} \cite{li2017diffusion} introduces diffusion graph convolution to describe the information diffusion process in spatial networks. \textsc{Dmstgcn} \cite{han2021dynamic} is based on \textsc{Stgcn} and learns the posterior graph for one day through back-propagation. \cite{lu2020spatiotemporal} exploits both spatial and semantic neighbors of of each node by constructing a dynamic weighted graph, and  the multi-head attention module is leveraged to capture the temporal dependencies among nodes. \textsc{Gman} \cite{zheng2020gman} uses spatial and temporal self-attention to capture dynamic spatial-temporal dependencies.  \textsc{Stgode} \cite{fang2021spatial} makes use of the ordinary differential equation (ODE) to model the spatial interactions of traffic flow. \textsc{ST-MetaNet}  is based on meta-learning and could conduct knowledge transfer across different time series, while the knowledge across cities is not considered \cite{pan2020spatio}.

Although impressive results have been achieved by works mentioned above, a few of them have discussed the transferability issue and cannot effectively utilize traffic data across cities. For example, \cite{lu2021learning} presents a multi-task learning framework for city heatmap-based traffic forecasting. \cite{mallick2021transfer} leverage a graph-partitioning method that decomposes a large highway network into smaller networks and uses a model trained on data-rich regions to predict traffic on unseen regions of the highway network.

\subsection{Transfer Learning with Deep Domain Adaptation}

The main challenge of transfer learning is to effectively reduce the discrepancy in data distributions across domains. Deep neural networks have the ability to extract transferable knowledge through representation learning methods \cite{yosinski2014transferable}. 
\cite{long2015learning} and \cite{long2018transferable} employ Maximum Mean Discrepancy (MMD) to improve the feature transferability and learn domain-invariant information. The conventional domain adaptation paradigm transfers knowledge from one source domain to one target domain. In contrast, multi-domain learning refers to a domain adaptation method in which multiple domains' data are incorporated in the training process \cite{nam2016learning,yang2014unified}. 

In recent years, adversarial learning has been explored for generative modeling in Generative Adversarial Networks (\textsc{Gan}s) \cite{goodfellow2014generative}. For example, Generative Multi-Adversarial Networks (\textsc{Gman}s) \cite{durugkar2016generative} extends \textsc{Gan}s to multiple discriminators including formidable adversary and forgiving teacher, which significantly eases model training and enhances distribution matching. In \cite{ganin2015unsupervised}, adversarial training is used to ensure that learned features in the shared space are indistinguishable to the discriminator and invariant to the shift between domains. \cite{pei2018multi} extends existing domain adversarial domain adaptation methods to multi-domain learning scenarios and proposed a multi-adversarial domain adaptation (\textsc{Mada}) approach to capture multi-mode structures to enable fine-grained alignment of different data distributions based on multiple domain discriminators.

\section{Preliminaries}\label{sec:preliminaries}

In this section, we first present definitions relevant to our work then rigorously formulate the transferable traffic forecasting problem.

\begin{definition}[Road Network $\mathcal{G}$]
A road network is represented as an undirected graph $\mathcal{G}=(V_{\mathcal{G}}, E_{\mathcal{G}}, A_{\mathcal{G}})$ to describe its topological structure. $V_{\mathcal{G}}$ is a set of nodes with $|V_{\mathcal{G}}|=N_{\mathcal{G}}$, $E_\mathcal{G}$ is a set of edges, and $A_{\mathcal{G}} \in \mathbb{R}^{N_{\mathcal{G}} \times N_{\mathcal{G}}}$ is the corresponding adjacency matrix of the road network. Particularly, we consider multiple road networks consisting of $|\mathcal{I}|$ source networks and one target network. $\mathcal{G}_{S_{i}}=(V_{\mathcal{G}_{S_i}}, E_{\mathcal{G}_{S_i}}, A_{\mathcal{G}_{S_i}})$ denotes the $i$th source road network ($i \in \mathcal{I}$), $\mathcal{G}_{T}=(V_{\mathcal{G}_{T}}, E_{\mathcal{G}_{T}}, A_{\mathcal{G}_{T}})$ denotes the target road network, and we have $|V_{\mathcal{G}_{S_i}}|=N_{\mathcal{G}_{S_i}}$, $|V_{\mathcal{G}_{T}}|=N_{\mathcal{G}_{T}}$. 
\end{definition}

\begin{definition}[Graph Signals ${X}$]
Let ${X_{\mathcal{G}}} \in \mathbb{R}^{N_\mathcal{G} \times N_{f}}$ denote the traffic state observed on $\mathcal{G}$ as a graph signal with node signals $X_v \in \mathbb{R}^{N_f}$ for $v \in V_{\mathcal{G}}$, where $N_{f}$ represents the number of features of each node ({\em e.g.}, flow, occupancy, speed).  Specifically, we use ${X}_{\mathcal{G}}^{(t)} \in \mathbb{R}^{N_{\mathcal{G}} \times N_{f}}$ to denote the observation on road network $\mathcal{G}$ at time $t$, and ${X}_{v}^{(t)} \in \mathbb{R}^{N_{f}}$ denotes the observation of node $v$ at time $t$, $\forall t\in \gamma$, where $\gamma$ is the study time period and $v \in V_{\mathcal{G}}$.
\end{definition}

We now define the transferable traffic forecasting problem.

\begin{definition}[Transferable traffic forecasting]

Given historical graph signals observed on both source and target domains as input, we can divide the transferable traffic forecasting problem into the pre-training and fine-tuning stages.  


%

In the pre-training stage, the forecasting task $\mathcal{T}_{S_i}$ maps $H^{\prime}$ historical node (graph) signals to future $H$ node (graph) signals on a source road network $\mathcal{G}_{S_{i}}$, for $v \in V_{\mathcal{G}_{S_i}}$:\begin{equation}\label{eq1}
    \left[{X}^{\left(t-H^{\prime}+1\right)}_{v}, \cdots, {X}^{(t)}_{v} ; \mathcal{G}_{S_i}   \right] \stackrel{\mathcal{T}_{S_i}(\cdot;\theta)}{\longrightarrow}\left[\widehat{{X}}^{(t+1)}_{v}, \cdots, \widehat{{X}}^{(t+H)}_{v}\right],
\end{equation}
where $\theta$ denotes the learned function parameters.

\begin{figure*}[t]
    \centering
    \includegraphics[width=0.85\textwidth]{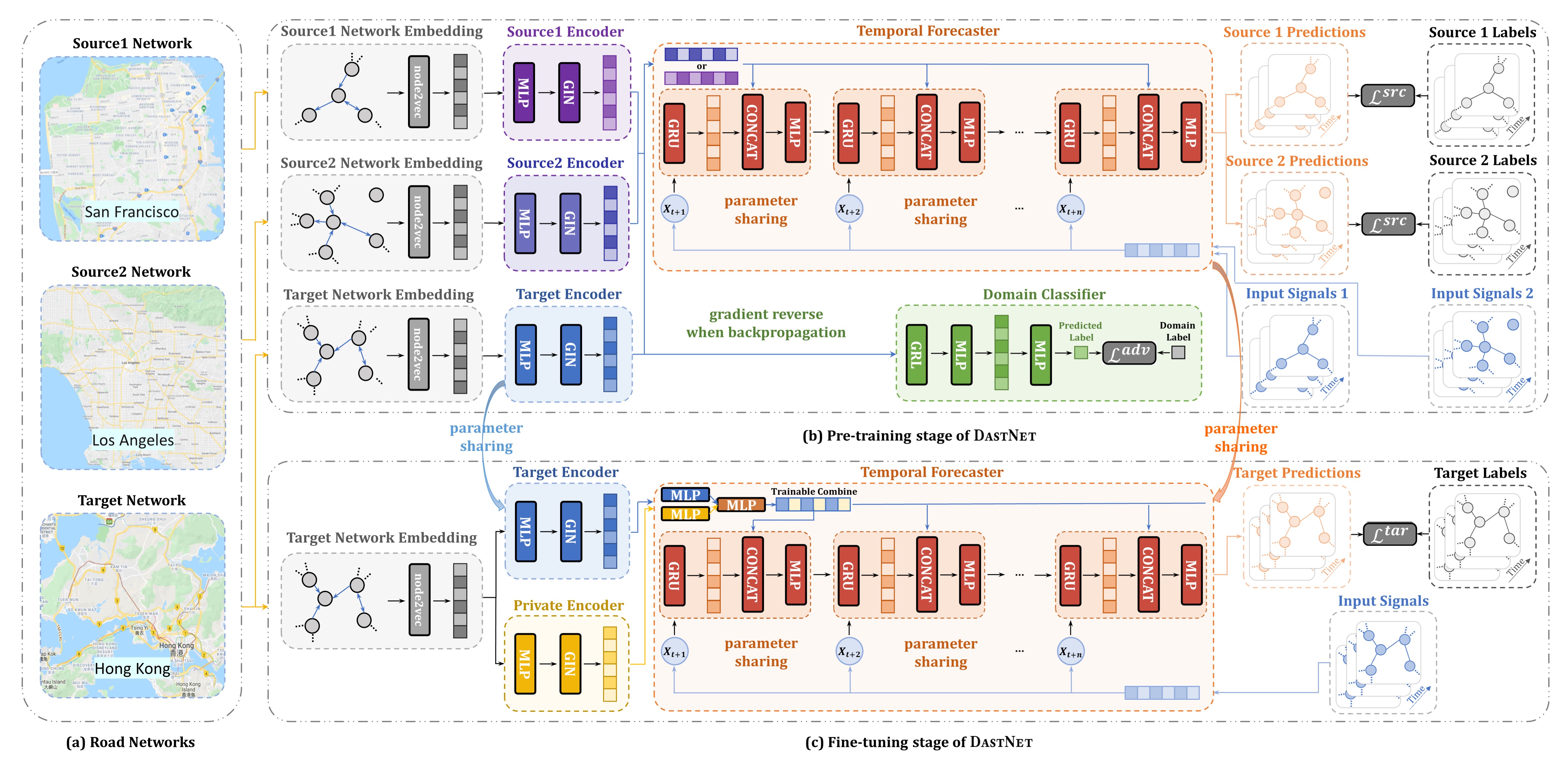}
    \caption{The proposed \textsc{DastNet} architecture.}
    \label{fig:model}
\end{figure*}

In the fine-tuning stage, to solve the forecasting task $\mathcal{T}_{T}$, the same function initialized with parameters $\theta$ shared from the pre-trained function is fine-tuned to predict graph signals on the target road network, for $v \in V_{\mathcal{G}_T}$:\begin{equation}\label{eq2}
    \left[{X}^{\left(t-H^{\prime}+1\right)}_{v}, \cdots, {X}^{(t)}_{v};  \mathcal{G}_{T}\right] \stackrel{\mathcal{T}_{T}\left(\cdot;\theta_*(\theta) \right)}{\longrightarrow}\left[\widehat{{X}}^{(t+1)}_{v}, \cdots, \widehat{{X}}^{(t+H)}_{v}\right],
\end{equation}
where $\theta_*(\theta)$ denotes the function parameters adjusted from $\theta$ to fit the target domain. 
\end{definition}
Note that the topology of $\mathcal{G}_{S_i}$ can be different from that of $\mathcal{G}_{T}$, and $\theta_*(\theta)$ represents the process of transferring the learned knowledge $\theta$ from $\mathcal{G}_{S_i}$ to the target domain $\mathcal{G}_{T}$. How to construct $\theta_*(\theta)$ to make it independent of network topology is the key research question in this study. To this end, the parameter sharing mechanism in the spatial \textsc{Gnn}s is utilized to construct $\theta_*(\theta)$ \cite{zhou2020graph}. For the following sections, we consider the study time period: $\gamma=\{(t-H^{\prime}+1),\cdots,(t+H)\}$. 


\section{Proposed Methodology}\label{sec:method}
In this section, we propose the domain adversarial spatial-temporal network (\textsc{DastNet}) to solve the transferable traffic forecasting problem. As shown in Figure \ref{fig:model}, \textsc{DastNet} is trained with two stages, and we use two source domains in the figure for illustration. We first perform pre-training through all the source domains in turn without revealing labels from the target domain. Then, the model is fine-tuned on the target domain. 
We will explain the pre-training stage and fine-tuning stage in detail, respectively.

\subsection{Stage 1: Pre-training on Source Domains}
In the pre-training stage, \textsc{DastNet} aims to learn domain-invariant knowledge that is helpful for forecasting tasks from multiple source domains. The learned knowledge can be transferred to improve the traffic forecasting tasks on the target domain. To this end, we design three major modules for \textsc{DastNet}: spatial encoder, temporal forecaster, and domain classifier.


The spatial encoder aims to consistently embed the spatial information of each node in different road networks. Mathematically, given a node $v$'s raw feature $\mathrm{e}_v \in \mathbb{R}^{D_{\mathrm{e}}}$, in which $D_\mathrm{e}$ is the dimension of raw features for each node, the spatial encoder maps it to a $D_{\mathrm{f}}$-dimensional node embedding $\mathrm{f}_v \in \mathbb{R}^{D_{\mathrm{f}}}$, i.e., $\mathrm{f}_v={M}_\mathrm{e}\left(\mathrm{e}_v; \theta_\mathrm{e}\right)$, where the parameters in this mapping $M_\mathrm{e}$ are denoted as $\theta_\mathrm{e}$. Note that the raw feature of a node can be obtained by a variety of methods ({\em e.g.}, POI information, GPS trajectories, geo-location information, and topological node representations).

Given a learned node embedding $\mathrm{f}_v$ for network $\mathcal{G}_i$, the temporal forecaster fulfils the forecasting task $\mathcal{T}_{S_i}$ presented in Equation \ref{eq1} by mapping historical node (graph) signals to the future node (graph) signals, which can be summarized by a mapping $(\widehat{X}_{v}^{(t+1)},\cdots,\widehat{X}_{v}^{(t+H)})=M_\mathrm{y}\left((X_{v}^{(t-H^{\prime}+1)},\cdots,X_{v}^{(t)}),\mathrm{f}_v;\theta_{\mathrm{y}}\right), \forall v \in V_{\mathcal{G}_{S_i}}$, and we denote the parameters of this mapping $M_\mathrm{y}$ as $\theta_\mathrm{y}$. 

Domain classifier takes node embedding $\mathrm{f}_v$ as input and maps it to the probability distribution vector $\widehat{\mathrm{d}}_v$ for domain labels, and we use notation $\mathrm{d}_v$ to denote the one-hot encoding of the actual domain label of $\mathrm{f}_v$. Note that the domain labels include all the source domains and the target domain. This mapping is represented as $\widehat{\mathrm{d}}_v={M}_\mathrm{d}\left(\mathrm{f}_v ; \theta_\mathrm{d}\right)$. We also want to make the node embedding $\mathrm{f}_v$ domain-invariant. That means, under the guidance of the domain classifier, we expect the learned node embedding $\mathrm{f}_v$ is independent of the domain label $\mathrm{d}_v$.

At the pre-training stage, we seek the parameters $(\theta_{\mathrm{e}}, \theta_{\mathrm{y}})$ of mappings $(M_\mathrm{e}, M_\mathrm{y})$ that minimize the loss of the temporal forecaster, while simultaneously seeking the parameters $\theta_{\mathrm{d}}$ of mapping $M_{\mathrm{d}}$ that maximize the loss of the domain classifier so that it cannot identify original domains of node embeddings learned from spatial encoders. 
Note that the target domain's node embedding is involved in the pre-training process to guide the target spatial encoder to learn domain-invariant node embeddings. Then we can define the loss function of the pre-training process as:
\begin{equation}\label{eq3}
    \begin{aligned}
\mathcal{L}(\cdot;\theta_{\mathrm{e}},\theta_{\mathrm{y}},\theta_{\mathrm{d}})=&\mathcal{L}^{src}\left(\cdot;\theta_{\mathrm{e}},\theta_{\mathrm{y}}\right)+\lambda\mathcal{L}^{adv}\left(\cdot;\theta_{\mathrm{e}},\theta_{\mathrm{d}}\right)\\
=&\mathcal{L}^{src}\left(M_{\mathrm{y}}\left(\cdot,M_{\mathrm{e}}\left(\mathrm{e}_v;\theta_{\mathrm{e}}\right);\theta_{\mathrm{y}}\right),\cdot\right) + \\ \lambda&\mathcal{L}^{adv}\left(M_{\mathrm{d}}\left(M_{\mathrm{e}}\left(\mathrm{e}_v;\theta_\mathrm{e}\right);\theta_\mathrm{d}\right),\mathrm{d}_v\right),
    \end{aligned}
\end{equation}

where $\lambda$ trades off the two losses.  
$\mathcal{L}^{src}(\cdot,\cdot)$ represents the prediction error on source domains and $\mathcal{L}^{adv}(\cdot,\cdot)$ is the adversarial loss for domain classification. 
Based on our objectives, we are seeking the parameters $\{\widehat{\theta}_\mathrm{e},\widehat{\theta}_\mathrm{y},\widehat{\theta}_\mathrm{d}\}$ that reach a saddle point of $\mathcal{L}$:
\begin{equation}\label{eq4}
\begin{aligned}
\left(\widehat{\theta}_\mathrm{e}, \widehat{\theta}_\mathrm{y}\right)&=\arg \min _{\theta_\mathrm{e}, \theta_\mathrm{y}} \mathcal{L}\left(\cdot;\theta_\mathrm{e}, \theta_\mathrm{y}, \widehat{\theta}_\mathrm{d}\right),\\
\widehat{\theta}_\mathrm{d}&=\arg \max _{\theta_\mathrm{d}} \mathcal{L}\left(\cdot;\widehat{\theta}_\mathrm{e}, \widehat{\theta}_\mathrm{y}, \theta_\mathrm{d}\right).
\end{aligned}
\end{equation}

Equation~\ref{eq4} essentially represents the min-max loss for \textsc{Gan}s, and the following sections will discuss the details of each component in the loss function.

\subsubsection{Spatial Encoder}
In traffic forecasting tasks, a successful transfer of trained \textsc{Gnn} models requires the adaptability of graph topology change between different road networks. To solve this issue, it is important to devise a graph embedding mechanism that can capture generalizable spatial information regardless of domains. To this end, we generate the raw feature $\mathrm{e}_v$ for a node $v$ by node2vec \cite{grover2016node2vec} as the input of the spatial encoder. Raw node features learned from the node2vec can reconstruct the ``similarity'' extracted from random walks since nodes are considered similar to each other if they tend to co-occur in these random walks. 
In addition to modeling the similarity between nodes, we also want to learn localized node features to identify the uniqueness of the local topology around nodes. 
\cite{xu2018powerful} proves that graph isomorphic network (\textsc{Gin}) layer is as powerful as the Weisfeiler-Lehman (WL) test \cite{weisfeiler1968reduction} for distinguishing different graph structures. Thus, we adopt \textsc{Gin} layers with mean aggregators proposed in \cite{xu2018powerful} as our spatial encoders. Mapping $\mathrm{f}_v={M}_\mathrm{e}\left(\mathrm{e}_v ; \theta_\mathrm{e}\right)$ can be specified by a $K$-layer \textsc{Gin} as follows:
\begin{equation}\label{eq5}
    \mathrm{f}_{v}^{(k)}=\mathrm{{MLP}_{gin}}^{(k)}\left(\left(1+\epsilon^{(k)}\right) \cdot \mathrm{f}_{v}^{(k-1)}+\sum_{u \in \mathcal{N}(v)} \frac{\mathrm{f}_{u}^{(k-1)}}{|\mathcal{N}(v)|} \right),
\end{equation}where $\mathrm{f}_{v}^{(0)}=\mathrm{e}_{v}$, $\mathcal{N}(v)$ denotes the neighborhoods of node $v$ and $\epsilon^{k}$ is a trainable parameter, $k=1, \cdots, K$, and $K$ is the total number of layers in \textsc{Gin}. The node $v$'s embedding can be obtained by $\mathrm{f}_v=\mathrm{f}_v^{(K)}$. We note that previous studies mainly use GPS trajectories to learn the location embedding \cite{wu2020learning, chen2021robust}, while this study utilizes graph topology and aggregate traffic data.

\subsubsection{Temporal Forecaster}
The learned node embedding $\mathrm{f}_v$ will be involved in the mapping ${M}_{\mathrm{y}}$ to predict future node signals. Now we will introduce our temporal forecaster, which aims to model the temporal dependencies of traffic data. Thus, we adapted the Gated Recurrent Units (\textsc{Gru}), which is a powerful \textsc{Rnn} variant \cite{DBLP:journals/corr/ChungGCB14, fu2016using}. In particular, we extend \textsc{Gru} by incorporating the learned node embedding $\mathrm{f}_v$ into its updating process.
To realize this, the learned node embedding $\mathrm{f}_v$ is concatenated with the hidden state of \textsc{Gru} (we denote the hidden state for node $v$ at time $\tau$ as $h_v^{(\tau)}$). Details of the mapping $M_{\mathrm{y}}$ is shown below: 
\begin{eqnarray}
u_{v}^{(\tau)}&=&\sigma\left(\Theta_{u} \left[{X}^{(\tau)}_{v}; h_{v}^{(\tau-1)}\right]+b_{u}\right),\label{eq6}\\
r_{v}^{(\tau)}&=&\sigma\left(\Theta_{r} \left[{X}^{(\tau)}_{v}; h_{v}^{(\tau-1)}\right]+b_{r}\right),\label{eq7}\\
c_{v}^{(\tau)}&=&\text{tanh}\left(\Theta_{c} \left[{X}^{(\tau)}_{v}; r_{v}^{(\tau)} \odot h_{v}^{(\tau-1)}\right]+b_{c}\right), \label{eq8}\\
h_{v}^{(\tau)}&=&\mathop{\mathrm{MLP^{(\tau)}_{gru}}(\mathrm{f}_{v}  ; (u_{v}^{(\tau)} \odot}_{\underset{\text{learned from spatial encoder}}{\downarrow}} h_{v}^{(\tau-1)}+(1-u_{v}^{(\tau)}) \odot c_{v}^{(\tau)})),\label{eq9}
\end{eqnarray}
where $u_v^{(\tau)}$, $r_v^{(\tau)}$, $c_v^{(\tau)}$ are update gate, reset gate and current memory content respectively. $\Theta_{u}$, $\Theta_{r}$, and $\Theta_{c}$ are parameter matrices, and $b_u$, $b_r$, and $b_c$ are bias terms.

The pre-training stage aims to minimize the error between the actual value and the predicted value. 
A single-layer perceptrons is designed as the output layer to map the temporal forecaster's output $h_v^{(\tau)}$ to the final prediction $\widehat{X}_{v}^{(\tau)}$. The source loss is represented by:
\begin{equation}\label{eq10}
    \mathcal{L}^{src}=\frac{1}{H}\sum_{\tau=t+1}^{t+H}\frac{1}{N_{\mathcal{G}_{S_i}}}\sum_{v \in V_{\mathcal{G}_{S_i}}}\left\|\widehat{X}^{(\tau)}_{v}-X^{(\tau)}_{v}\right\|_{1}.
\end{equation}

\subsubsection{Domain Classifier}
The difference between domains is the main obstacle in transfer learning. In the traffic forecasting problem, the primary domain difference that leads to the model's inability to conduct transfer learning between different domains is the spatial discrepancy. Thus, spatial encoders are involved in learning domain-invariant node embeddings for both source networks and the target network in the pre-training process.

To achieve this goal, we involve a Gradient Reversal Layer (GRL) \cite{ganin2015unsupervised} and a domain classifier trained to distinguish the original domain of node embedding. The GRL has no parameters and acts as an identity transform during forward propagation. During the backpropagation, GRL takes the subsequent level's gradient, and passes its negative value to the preceding layer. 
In the domain classifier, given an input node embedding $\mathrm{f}_v$, $\theta_{\mathrm{d}}$ is optimized to predict the correct domain label, and $\theta_{\mathrm{e}}$ is trained to maximize the domain classification loss. Based on the mapping $\widehat{\mathrm{d}}_v={M}_\mathrm{d}\left(\mathrm{f}_v  ; \theta_\mathrm{d}\right) = \operatorname{Softmax} \left( \operatorname{MLP}_{\mathrm{d}}\left(\mathrm{f}_{v}\right)\right)$, $\mathcal{L}^{adv}$ is defined as:

\begin{equation}\label{eq11}
{\mathcal{L}}^{a d v}=\sum_{V \in V_{\text {all }}}-\frac{1}{|V|} \sum_{v \in V}  \langle \mathrm{d}_{v}, \log \left(\operatorname{Softmax}\left(\operatorname{MLP}_{\mathrm{d}}\left(\mathrm{f}_{v}\right)\right)\right) \rangle,
\end{equation}
where $V_{all}=V_{\mathcal{G}_{S_1}} \cup V_{\mathcal{G}_{S_2}} \cup V_{\mathcal{G}_{T}}$, and the output of $\mathrm{MLP}_\mathrm{d}(\mathrm{f}_v)$ is fed into the softmax, which computes the possibility vector of node $v$ belonging to each domain.

By using the domain adversarial learning, we expect to learn the ``forecasting-related knowledge'' that is independent of time, traffic conditions, and traffic conditions. The idea of spatial encoder is also inspired by the concept of land use regression (LUR) \cite{steininger2020maplur}, which is originated from geographical science. The key idea is that the location itself contains massive information for estimating traffic, pollution, human activities, and so on. If we can properly extract such information, the performance of location-related tasks can be improved.

\subsection{Stage 2: Fine-tuning on the Target Domain}

The objective of the fine-tuning stage is to utilize the knowledge gained from the pre-training stage to further improve forecasting performance on the target domain. 
Specifically, we adopt the parameter sharing mechanism in \cite{pan2009survey}: the parameters of the target spatial encoder and the temporal forecaster in the fine-tuning stage are initialized with the parameters trained in the pre-training stage.

Moreover, we involve a private spatial encoder combined with the pre-trained target spatial encoder to explore both domain-invariant and domain-specific node embeddings. Mathematically, given a raw node feature $\mathrm{e}_v$, the private spatial encoder maps it to a domain-specific node embedding $\tilde{\mathrm{f}}_v$, this process is represented as $\tilde{\mathrm{f}}_v=\tilde{M}_{\mathrm{e}}(\mathrm{e}_v;\tilde{\theta}_{\mathrm{e}})$, where $\tilde{M}_{\mathrm{e}}$ has the same structure as $M_{\mathrm{e}}$ and the parameter $\tilde{\theta}_{\mathrm{e}}$ is randomly initialized. The pre-trained target spatial encoder maps the raw node feature $\mathrm{e}_v$ to a domain-invariant node embedding $\mathrm{f}_v$, i.e., $\mathrm{f}_v=M_{\mathrm{e}}(\mathrm{e}_v;\theta_{\mathrm{e}*}(\theta_{\mathrm{e}}))$, where $\theta_{\mathrm{e}*}(\theta_{\mathrm{e}})$ means that $\theta_{\mathrm{e}*}$ is initialized with the trained parameter $\theta_{\mathrm{e}}$ in the pre-training stage. Note that $\tilde{M}_{\mathrm{e}}$ and $M_{\mathrm{e}}$  are of the same structure, and the process to generate $\tilde{\mathrm{f}}_v$ and $\mathrm{f}_v$ is the same as in Equation \ref{eq5}.

Before being incorporated into the pre-trained temporal forecaster,   $\tilde{\mathrm{f}}_v$ and $\mathrm{f}_v$ are combined by $\mathrm{MLP}$ layers to learn the combined node embedding $\mathrm{f}_v^{tar}$ of the target domain:
\begin{equation}\label{eq12}
    \mathrm{f}_v^{tar}=\mathrm{{MLP}_{cmb}}\left(\mathrm{{MLP}_{pre}}(\mathrm{f}_v) + \mathrm{{MLP}_{pri}}(\tilde{\mathrm{f}}_v)\right),
\end{equation}
then given node signal $X_{v}^{(\tau)}(v \in V_{\mathcal{G}_T})$ at time $\tau$ and $\mathrm{f}_v^{\mathrm{tar}}$ as input, $\widehat{X}_v^{(\tau)}$ is computed based on Equation \ref{eq6}, \ref{eq7}, \ref{eq8}, and \ref{eq9}. We denote the target loss at the fine-tuning stage as:
\begin{equation}\label{eq13}
    \mathcal{L}^{tar}=\frac{1}{H}\sum_{\tau=t+1}^{t+H}\frac{1}{N_{\mathcal{G}_T}}\sum_{v \in V_{\mathcal{G}_{T}}}\left\|\widehat{X}^{(\tau)}_{v}-X^{(\tau)}_{v}\right\|_{1}.
\end{equation}

\section{Experiments}\label{sec:exp}
We first validate the performance of \textsc{DastNet} using benchmark datasets, and then \textsc{DastNet} is experimentally deployed with the newly collected data in Hong Kong.
\subsection{Offline Validation with Benchmark Datasets}




\begin{figure}[h]
    \centering
    \includegraphics[width=.32\textwidth]{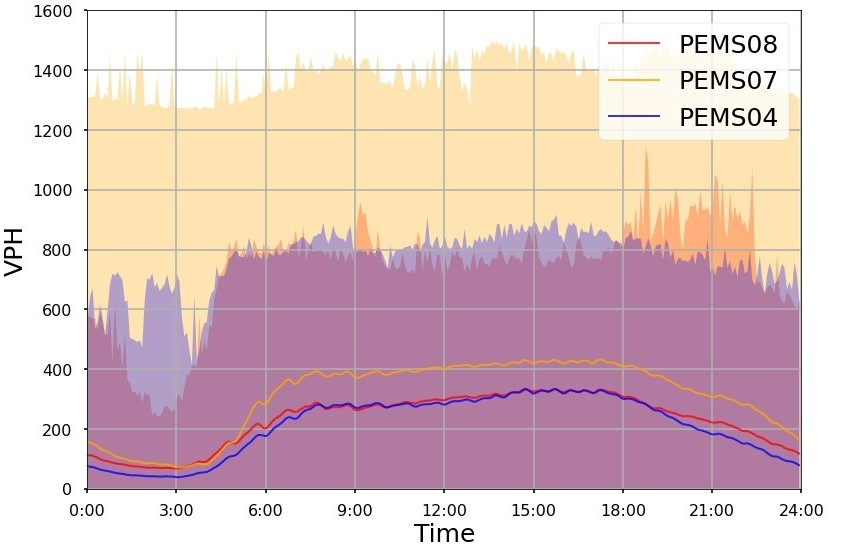}
    \caption{Within-day traffic flow distributions.}
    \label{fig:distributions}
\end{figure}

\renewcommand{\arraystretch}{0.85}
\begin{table*}[t]
\caption{Performance comparison of different methods. (mean $\pm$ std)} 
\centering 

\resizebox{0.84\textwidth}{!}{
\Large
\begin{tabular}{c||ccc|ccc|ccc}
\arrayrulecolor{black!30}\midrule
\hline
\multirow{2}{*}{PEMS04} & \multicolumn{3}{c|}{15min}                                                         & \multicolumn{3}{c|}{30min}                                                         & \multicolumn{3}{c}{60min}                                                         \\  \cline{2-10} 
                       & \multicolumn{1}{c}{MAE} & \multicolumn{1}{c}{RMSE} & \multicolumn{1}{c|}{MAPE(\%)} & \multicolumn{1}{c}{MAE} & \multicolumn{1}{c}{RMSE} & \multicolumn{1}{c|}{MAPE(\%)} & \multicolumn{1}{c}{MAE} & \multicolumn{1}{c}{RMSE} & \multicolumn{1}{c}{MAPE(\%)} \\ \hline
\textsc{Ha}            & 28.36±0.00  & 40.55±0.00  & 20.14±0.00  & 31.75±0.00  & 45.14±0.00  & 22.84±0.00  & 38.52±0.00  & 54.45±0.00  & 28.48±0.00 \\
\textsc{Svr}           & 21.21±0.05  & 29.68±0.07  & 16.05±0.14  & 23.90±0.04  & 33.51±0.02  & 18.74±0.40  & 29.24±0.14  & 41.14±0.10  & 23.46±0.73 \\
\textsc{Gru}           & 20.96±0.29  & 31.08±0.20  & 14.78±1.86  & 22.71±0.21  & 33.77±0.19  & 16.54±1.73  & 26.25±0.28  & 38.87±0.32  & 18.66±1.95 \\
\textsc{Gcn}           & 48.65±0.04  & 68.89±0.06  & 40.53±0.88  & 49.49±0.05  & 69.97±0.06  & 41.42±0.78  & 51.63±0.06  & 72.65±0.07  & 44.03±0.49 \\
\textsc{Tgcn}          & 24.09±1.35  & 34.31±1.59  & 18.26±1.38  & 25.22±0.96  & 36.09±1.22  & 19.34±1.07  & 27.16±0.65  & 38.76±0.94  & 20.84±0.37 \\
\textsc{Stgcn}         & 27.03±1.30  & 38.26±1.35  & 25.16±4.33  & 27.91±0.88  & 39.65±0.78  & 25.33±5.06  & 35.55±2.43  & 49.12±4.01  & 37.74±5.15 \\
\textsc{Dcrnn}         & 23.73±0.62  & 34.27±0.71  & 18.84±0.75  & 26.68±0.94  & 37.63±1.00  & 21.39±1.90  & 33.79±1.77  & 46.70±1.91  & 29.68±1.76 \\
\textsc{Agcrn}         & 24.58±0.35  & 42.30±0.30  & 14.93±0.13  & 26.53±0.20  & 48.05±0.52  & 15.30±0.36  & 30.06±0.29  & 52.19±0.55  & 16.67±0.07 \\
\textsc{Stgode}        & 20.73±0.04  & 31.97±0.06  & 15.79±0.22  & 23.14±0.08   & 35.55±0.23  & 17.66±0.16  & 27.24±0.08  & 41.05±0.10  & 23.86±0.38 \\
\arrayrulecolor{black!30}\midrule
\textsc{Temporal Forecaster}     & 20.70±0.60  & 30.80±0.46  & 14.72±1.91  & 22.22±0.15  & 33.19±0.13  & 15.53±0.76  & 25.88±0.09  & 38.33±0.12  & 17.84±1.09 \\
\textsc{Target Only}   & 19.81±0.06  & 29.77±0.03  & 13.95±0.47  & 21.55±0.09  & 32.26±0.13  & 14.83±0.21  & 24.59±0.13  & 36.31±0.15  & 17.45±0.39 \\
\textsc{DastNet w/o Da}         & 19.65±0.11  & 29.52±0.14  & 13.53±0.35  & 21.57±0.41  & 32.26±0.76  & 15.09±0.54  & 23.84±0.10  & 35.21±0.14  & 17.03±0.44 \\
\textsc{DastNet w/o Pri}        & 19.35± 0.09 & 29.05±0.15  & 13.54±0.24  & 21.00±0.54  & 31.40±0.87  & 14.61±0.31  & 22.96±0.38  & 34.02±0.54  & 16.51±0.58 \\
\textsc{DastNet}                  & \textbf{19.25±0.03}  & \textbf{28.91±0.05}  & \textbf{13.30±0.22}  & \textbf{20.67±0.07}  & \textbf{30.78±0.04}  & \textbf{14.56±0.31}  & \textbf{22.82±0.08}  & \textbf{33.77±0.13}  & \textbf{16.10±0.18} \\ \hline\hline
\multirow{2}{*}{PEMS07} & \multicolumn{3}{c|}{15min}                                                         & \multicolumn{3}{c|}{30min}                                                         & \multicolumn{3}{c}{60min}                                                         \\ \cline{2-10}
                       & \multicolumn{1}{c}{MAE} & \multicolumn{1}{c}{RMSE} & \multicolumn{1}{c|}{MAPE(\%)} & \multicolumn{1}{c}{MAE} & \multicolumn{1}{c}{RMSE} & \multicolumn{1}{c|}{MAPE(\%)} & \multicolumn{1}{c}{MAE} & \multicolumn{1}{c}{RMSE} & \multicolumn{1}{c}{MAPE(\%)} \\ \hline
\textsc{Ha}                     & 32.85±0.00  & 46.56±0.00  & 15.10±0.00  & 37.09±0.00  & 52.38±0.00  & 17.26±0.00  & 45.43±0.00  & 63.93±0.00  & 21.66±0.00 \\
\textsc{Svr}                    & 23.36±0.38  & 32.30±0.28  & 14.97±1.41  & 27.33±0.30  & 37.60±0.22  & 19.23±0.89  & 36.90±0.98  & 49.13±0.77  & 33.50±2.83 \\
\textsc{Gru}                    & 23.77±0.49  & 34.49±0.52  & 11.21±0.66  & 25.31±0.37  & 37.85±0.38  & 12.87±2.08  & 29.39±0.25  & 43.89±0.35  & 13.26±0.37 \\
\textsc{Gcn}                    & 50.81±0.56  & 71.67±0.50  & 36.47±1.57  & 51.94±0.24  & 73.18±0.30  & 39.10±1.26  & 55.09±0.07  & 77.15±0.10  & 41.46±0.42 \\
\textsc{Tgcn}                   & 30.18±0.41  & 42.11±0.56  & 15.74±0.99  & 30.84±2.77  & 43.58±3.37  & 15.19±1.59  & 33.25±1.45  & 47.24±1.82  & 16.58±1.04 \\
\textsc{Stgcn}                  & 34.14±6.13  & 48.58±7.32  & 19.67±6.38  & 39.50±2.76  & 43.58±3.37  & 15.09±1.59  & 43.45±2.50  & 60.67±3.23  & 27.57±1.36 \\
\textsc{Dcrnn}                  & 26.66±1.23  & 37.66±1.39  & 16.68±1.31  & 31.06±1.39  & 43.38±1.75  & 19.94±2.48  & 51.09±6.82  & 66.26±7.42  & 48.29±17.74 \\
\textsc{Agcrn}                  & 35.16±0.23  & 64.08±0.45  & 11.88±0.12  & 35.10±0.25  & 63.78±0.44  & 11.98±0.14  & 39.00±1.74  & 68.44±0.41  & 13.98±0.04 \\
\textsc{Stgode}                 & 22.30±0.13  & 33.89±0.14  & 10.92±0.20  & 26.02±0.18  & 38.52±0.14  & 14.23±0.57  & 30.87±0.43  & 45.27±0.25  & 17.21±1.57 \\
\arrayrulecolor{black!30}\midrule
\textsc{Temporal Forecaster}    & 23.11±0.54  & 34.07±0.38  & 10.97±1.25  & 24.70±0.20  & 37.13±0.22  & 10.98±0.58  & 28.55±0.18  & 42.72±0.22  & 12.67±0.17 \\ 
\textsc{Target Only}            & 21.71±0.13  & 32.93±0.22  &  9.41±0.11  & 24.61±1.00  & 37.15±1.46  & 10.80±0.69  & 28.88±0.65  & 43.13±0.98  & 13.18±0.77 \\
\textsc{DastNet  w/o Da}        & 21.80±0.26  & 33.09±0.44  &  9.45±0.18  & 24.52±0.55  & 37.05±0.94  & 10.77±0.42  & 28.61±0.56  & 42.88±0.91  & 12.74±0.42 \\
\textsc{DastNet  w/o Pri}       & 21.23±0.14  & 32.28±0.24  & 9.20± 0.15  & 23.85±0.47  & 36.10±0.71  & 10.51±0.22  & 28.37±1.06  & 42.51±1.64  & 12.74±0.50   \\
\textsc{DastNet}                & \textbf{20.91±0.03}  & \textbf{31.85±0.05}  &  \textbf{8.95±0.13}  & \textbf{22.96±0.10}  & \textbf{34.80±0.11}  &  \textbf{9.87±0.19}  & \textbf{26.88±0.28}  & \textbf{40.12±0.29}  & \textbf{11.75±0.33} \\ \hline\hline
\multirow{2}{*}{PEMS08} 
                       & \multicolumn{3}{c|}{15min}                                                         & \multicolumn{3}{c|}{30min}                                                         & \multicolumn{3}{c}{60min}                                                         \\ \cline{2-10}
                       & \multicolumn{1}{c}{MAE} & \multicolumn{1}{c}{RMSE} & \multicolumn{1}{c|}{MAPE(\%)} & \multicolumn{1}{c}{MAE} & \multicolumn{1}{c}{RMSE} & \multicolumn{1}{c|}{MAPE(\%)} & \multicolumn{1}{c}{MAE} & \multicolumn{1}{c}{RMSE} & \multicolumn{1}{c}{MAPE(\%)} \\ \hline
\textsc{Ha}                     & 23.12±0.00  & 33.03±0.00  & 14.61±0.00  & 26.12±0.00  & 37.16±0.00  & 16.55±0.00  & 32.15±0.00  & 45.41±0.00  & 20.60±0.00 \\
\textsc{Svr}                    & 37.63±2.42  & 46.59±2.56  & 20.79±1.47  & 45.79±2.59  & 56.16±2.70  & 24.29±1.02  & 66.91±3.82  & 79.72±4.07  & 33.20±1.86 \\
\textsc{Gru}                    & 16.69±0.40  & 24.72±0.41  & 11.05±0.93  & 18.89±0.67  & 28.14±0.65  & 13.45±3.18  & 20.94±0.24  & 31.32±0.19  & 15.20±0.94 \\
\textsc{Gcn}                    & 64.63±0.08  & 87.30±0.10  & 90.32±1.83  & 65.09±0.06  & 87.87±0.08  & 91.64±1.12  & 66.24±0.11  & 89.21±0.10  & 94.01±1.93 \\
\textsc{Tgcn}                   & 20.65±0.96  & 28.77±1.13  & 15.06±1.20  & 21.60±1.44  & 30.40±1.78  & 15.97±2.42  & 24.33±2.51  & 34.20±3.14  & 17.91±4.77 \\
\textsc{Stgcn}                  & 25.90±1.60  & 35.58±1.98  & 18.91±2.35  & 26.20±1.75  & 36.52±2.34  & 17.73±0.74  & 31.89±4.23  & 43.94±5.56  & 20.99±2.41 \\
\textsc{Dcrnn}                  & 20.61±0.97  & 29.03±1.08  & 20.36±1.62  & 23.23±1.24  & 32.76±1.44  & 24.53±2.77  & 39.14±7.12  & 51.97±8.41  & 47.62±19.08 \\
\textsc{Agcrn}                  & 18.50±0.16  & 30.76±0.30  & 10.77±0.09  & 19.45±0.12  & 32.34±0.23  & 11.30±0.09  & 23.44±0.13  & 37.55±0.19  & 13.71±0.07 \\
\textsc{Stgode}                 & 20.42±0.69  & 37.92±3.06  & 17.82±1.08  & 23.41±0.48  & 36.41±2.89  & 21.00±2.18  & 26.86±0.28  & 39.85±0.57  & 24.43±0.02 \\
\arrayrulecolor{black!30}\midrule
\textsc{Temporal Forecaster}    & 15.99±0.10  & 23.95±0.11  &  9.93±0.45  & 17.77±0.40  & 26.56±0.31  & 12.08±1.75  & 20.03±0.33  & 29.86±0.21  & 14.80±2.10 \\ 
\textsc{Target Only}            & 16.50±0.12  & 24.58±0.12  & 11.07±0.16  & 17.95±1.04  & 26.63±1.24  & 11.90±2.09  & 19.69±0.33  & 29.37±0.40  & 12.48±0.37 \\
\textsc{DastNet  w/o Da}        & 16.51±0.30  & 24.50±0.38  & 10.55±0.99  & 17.58±0.81  & 26.31±1.21  & 11.22±0.76  & 19.37±0.46  & 28.87±0.54  & 11.95±0.36 \\
\textsc{DastNet  w/o Pri}       & 15.75±0.25  & 23.60±0.41  & 10.00±0.22  & 16.87±0.38  & 25.38±0.68  & 10.55±0.14  & 18.90±0.20  & 28.28±0.20  & 12.52±0.64 \\
\textsc{DastNet}                & \textbf{15.26±0.18}  & \textbf{22.70±0.17}  & \textbf{9.64±0.37}  & \textbf{16.41±0.34}  & \textbf{24.57±0.39}  & \textbf{10.46±0.31}  & \textbf{18.84±0.12}  & \textbf{28.06±0.17}  & \textbf{11.72±0.29} \\ \hline\hline
\end{tabular}
}
\label{table:mainresults}
\end{table*}

We evaluate the performance of \textsc{DastNet} on three real-world datasets, PEMS04, PEMS07, PEMS08, which are collected from the Caltrans Performance Measurement System (PEMS) \cite{caltrans} every 30 seconds. There are three kinds of traffic measurements in the raw data: speed, flow, and occupancy. In this study, we forecast the traffic flow for evaluation purposes and it is aggregated to 5-minute intervals, which means there are 12 time intervals for each hour and 288 time intervals for each day. The unit of traffic flow is veh/hour (vph).
The within-day traffic flow distributions are shown in Figure \ref{fig:distributions}. One can see that flow distributions vary significantly over the day for different datasets, and hence domain adaption is necessary.

The road network for each dataset are constructed according to actual road networks, and we defined the adjacency matrix based on connectivity. Mathematically, $A_{i, j} = \begin{cases}1, & \text { if } v_{i} \text { connects to } v_{j} \\ 0, & \text { otherwise }\end{cases}$, where $v_i$ denotes node $i$ in the road network. Moreover, we normalize the graph signals by the following formula: ${X}=\frac{{X}-\operatorname{mean}({X})}{\operatorname{std}({X})}$, where function $\operatorname{mean}$ and function $\operatorname{std}$ calculate the mean value and the standard deviation of historical traffic data respectively.

\subsubsection{Baseline Methods}
\begin{itemize}[leftmargin=*]
    \item \textsc{Ha} \cite{liu2004summary}: Historical Average method uses average value of historical traffic flow data as the prediction of future traffic flow.
    \item \textsc{Svr} \cite{smola2004tutorial}: Support Vector Regression adopts support vector machines to solve regression tasks.
    \item \textsc{Gru} \cite{cho2014properties}: Gated Recurrent Unit (\textsc{Gru}) is a well-known variant of \textsc{Rnn} which is powerful at capturing temporal dependencies.
    \item \textsc{Gcn} \cite{kipf2016semi}: Graph Convolutional Network can handle arbitrary graph-structured data and has been proved to be powerful at capturing spatial dependencies.
    \item \textsc{Tgcn} \cite{zhao2019t}: Temporal Graph Convolutional Network performs stably well for short-term traffic forecasting tasks.
    \item \textsc{Stgcn} \cite{yu2017spatio}: Spatial-Temporal Graph Convolutional Network uses \textsc{ChebNet} and 2D convolutions for traffic prediction.
    \item \textsc{Dcrnn} \cite{li2017diffusion}: Diffusion Convolutional Recurrent Neural Network combines \textsc{Gnn} and \textsc{Rnn} with diffusion convolutions.
    \item \textsc{Agcrn} \cite{bai2020adaptive}: Adaptive Graph Convolutional Recurrent Network learns node-specific patterns through node adaptive parameter learning and data-adaptive graph generation.
    \item \textsc{Stgode} \cite{fang2021spatial}: Spatial-Temporal Graph Ordinary Differential Equation Network capture spatial-temporal dynamics through a tensor-based ODE.
\end{itemize}
To demonstrate the effectiveness of each key module in \textsc{DastNet}, we compare with some variants of \textsc{DastNet} as follows:

\begin{itemize}[leftmargin=*]
    \item \textsc{Temporal Forecaster}: \textsc{DastNet} with only the \textsc{Temporal Forecaster} component. This method only uses graph signals as input for pre-training and fine-tuning.
    \item \textsc{Target Only}: \textsc{DastNet} without training at the pre-training stage. The comparison with this baseline method demonstrate the merits of training on other data sources.
    \item \textsc{DastNet  w/o Da}: \textsc{DastNet} without the adversarial domain adaptation (domain classifier).  The comparison with this baseline method demonstrate the merits of domain-invariant features.
    \item \textsc{DastNet  w/o Pri}: \textsc{DastNet} without the private encoder at the fine-tuning stage.
\end{itemize}

Above variants' other settings are the same as \textsc{DastNet}.

\subsubsection{Experimental Settings}
To simulate the lack of data, for each dataset, we randomly select ten consecutive days' traffic flow data from the original training set as our training set, and the validation/testing sets are the same as \cite{li2021dynamic}.
We use one-hour historical traffic flow data for training and forecasting traffic flow in the next 15, 30, and 60 minutes (horizon=3, 6, and 12, respectively). For one dataset $\mathcal{D}$, \textsc{DastNet}-related methods and \textsc{Trans Gru} are pre-trained on the other two datasets, and fine-tuned on $\mathcal{D}$. Other methods are trained on $\mathcal{D}$. All experiments are repeated 5 times. Other hyper-parameters are determined based on the validation set. 
We implement our framework based on Pytorch \cite{paszke2019pytorch} on a virtual workstation with two 11G memory Nvidia GeForce RTX 2080Ti GPUs. To suppress noise from the domain classifier at the early stages of the pre-training procedure, instead of fixing the adversarial domain adaptation loss factor $\mathcal{F}$. We gradually change it from 0 to 1: $\mathcal{F}=\frac{2}{1+\exp (-\eta \cdot \mathcal{P})}-1$, where $\mathcal{P}= \frac{\text{current step}}{\text{total steps}}$, $\eta$ was set to 10 in all experiments. We select the SGDM optimizer for stability and set the maximum epochs for fine-tuning stage to 2000 and set K of \textsc{Gin} encoders as 1 and 64 as the dimension of node embedding. For all model we set 64 as the batch size. 
For node2vec settings, we set $p=q=1$, 
and each source node conduct 200 walks with 8 as the walk length and 64 as the embedding dimension.



Table \ref{table:mainresults} shows the performance comparison of different methods for traffic flow forecasting. Let $x_{i} \in {X}$ denote the ground truth and $\hat{x}_{i}$ represent the predicted values, and $\Omega$ denotes the set of training samples' indices. The performance of all methods are evaluated based on (1) Mean Absolute Error ($\operatorname{MAE}(x, \hat{x})=\frac{1}{|\Omega|} \sum_{i \in \Omega}\left|x_{i}-\hat{x}_{i}\right|$), which is a fundamental metric to reflect the actual situation of the prediction accuracy. (2) Root Mean Squared Error ($\operatorname{RMSE}(x, \hat{x})=\sqrt{\frac{1}{|\Omega|} \sum_{i \in \Omega}\left(x_{i}-\hat{x}_{i}\right)^{2}}$), which is more sensitive to abnormal values. (3) Mean Absolute Percentage Error ($\operatorname{MAPE}(x, \hat{x})=\frac{1}{|\Omega|} \sum_{i \in \Omega}\left|\frac{x_{i}-\hat{x}_{i}}{x_{i}}\right|$). It can be seen that \textsc{DastNet} achieves the state-of-the-art forecasting performance on the three datasets for all evaluation metrics and all prediction horizons. Traditional statistical methods like \textsc{Ha} and \textsc{Svr} are less powerful compared to deep learning methods such as \textsc{Gru}. 
The performance of \textsc{Gcn} is low, as it overlooks the temporal patterns of the data. 
\textsc{DastNet} outperforms existing spatial-temporal models like \textsc{Tgcn}, \textsc{Stgcn}, \textsc{Dcrnn}, \textsc{Agcrn} and the state-of-the-art method \textsc{Stgode}. \textsc{DastNet} achieves approximately 9.4\%, 8.6\% and 10.9\% improvements compared to the best baseline method in MAE, RMSE, MAPE, respectively. Table \ref{table:ImpvTO} summarize the improvements of our methods, where "-" denotes no improvements.. 


\renewcommand{\arraystretch}{0.86}
\begin{table}[h]
\setlength{\tabcolsep}{1pt}
\caption{Comparison between 1) \textsc{Gru} and the \textsc{Target Only} (Upper); 2) \textsc{DastNet} and the best baseline (Lower).}
\centering
\resizebox{0.4\textwidth}{!}{
\begin{tabular}{c||ccc|ccc|ccc}
\hline \hline 
\multirow{2}{*}{Impv.} & \multicolumn{3}{c|}{15min} & \multicolumn{3}{c|}{30min} & \multicolumn{3}{c}{60min} \\ \cline{2-10}
                    & MAE      & RMSE    & MAPE   & MAE    & RMSE    & MAPE   & MAE    & RMSE    & MAPE   \\ \hline
04                  & 5.4\%    & 4.2\%  & 6.2\% & 5.1\%       & 4.4\%        & 10\%       & 6.3\%       & 6.5\%        & 6.5\%       \\ \hline
07                  & 8.6\%    & 4.5\%  & 16\%  & 2.8\%       & 1.8\%        & 16\%       & 1.7\%       & 1.7\%        & 0.6\%       \\ \hline
08                  & 1.1\%    & 0.6\%  & -     & 5.0\%       & 5.4\%        & 11.5\%     & 6.0\%       & 6.2\%        & 18.0\%      \\
\hline \hline
\multirow{2}{*}{Impv.} & \multicolumn{3}{c|}{15min} & \multicolumn{3}{c|}{30min} & \multicolumn{3}{c}{60min} \\ \cline{2-10}
                    & MAE    & RMSE    & MAPE   & MAE    & RMSE    & MAPE   & MAE    & RMSE    & MAPE   \\ \hline
04                  & 7.1\%  & 2.6\%   & 10.9\%  & 9.0\%  & 8.2\%  & 4.8\%        & 13.1\%      & 12.9\%       & 3.4\%       \\ \hline
07                  & 6.2\%  & 6.5\%   & 18.0\%  & 9.3\%  & 7.4\%  & 17.6\%       & 8.5\%       & 8.6\%        & 11.4\%       \\ \hline
08                  & 8.6\%  & 8.2\%   & 10.5\%  & 13.1\%  & 12.7\%  & 7.4\%       & 10.0\%       & 10.4\%        & 14.5\%      \\\hline \hline 
\end{tabular}
}
\label{table:ImpvTO}
\end{table}

\noindent\textbf{Ablation Study.}
From Table~\ref{table:mainresults}, MAE, RMSE and MAPE of the \textsc{Target Only} are reduced by approximately 4.7\%, 7\% and 10.6\% compared to \textsc{Gru} (see Table \ref{table:ImpvTO}), which demonstrates that the temporal forecaster outperforms \textsc{Gru} due to the incorporation of the learned node embedding. The accuracy of \textsc{DastNet} is superior to \textsc{Target Only}, \textsc{DastNet w/o Da}, \textsc{Temporal Forecaster} and \textsc{DastNet w/o Pri}, which shows the effectiveness of pre-training, adversarial domain adaptation, spatial encoders and the private encoder. 
Interestingly, the difference between the results of the \textsc{DastNet} and \textsc{DastNet w/o Pri} on PEMS07 is generally larger than that on dataset PEMS04 and PEMS08. According to Figure \ref{fig:distributions}, we know that the data distribution of PEMS04 and PEMS08 datasets are similar, while the data distribution of PEMS07 is more different from that of PEMS04 and PEMS08. This reflects differences between spatial domains and further implies that our private encoder can capture the specific domain information and supplement the information learned from the domain adaptation. 


\begin{figure}[h]
    \centering
    \includegraphics[width=.46\textwidth]{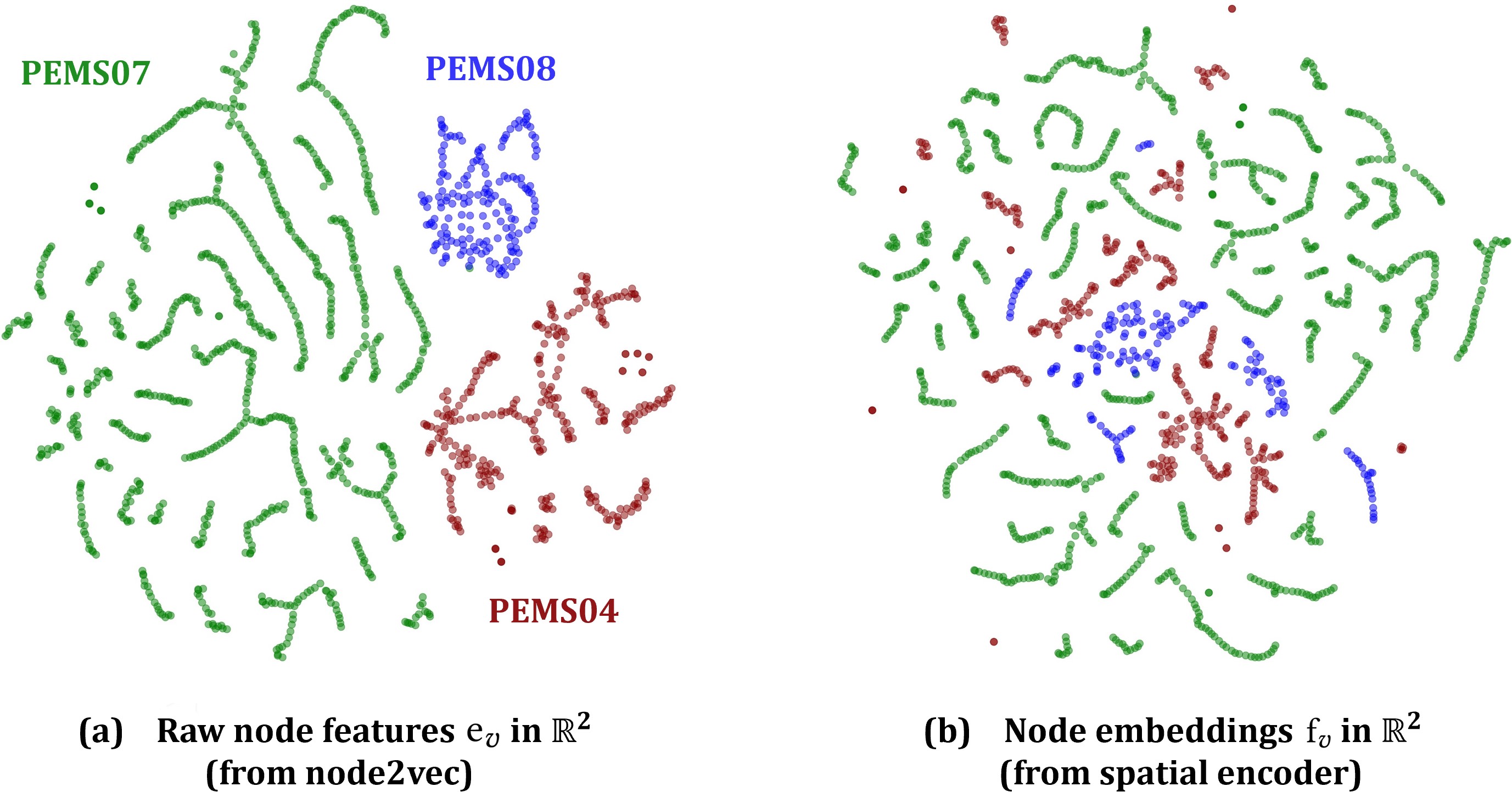}
    \caption{Visualization of $\mathrm{e}_v$ and $\mathrm{f}_v$ by t-SNE.}
    \label{fig:encode}
\end{figure}

\noindent\textbf{Effects of Domain Adaptation.}
To demonstrate the effectiveness of the proposed adversarial domain adaptation module, we visualize the raw feature of the node $\mathrm{e}_v$ (generated from node2vec) and the learned node embedding $\mathrm{f}_v$ (generated from spatial encoders) in Figure \ref{fig:encode} using t-SNE \cite{van2013barnes}. As illustrated, node2vec learns graph connectivity for each specific graph, and hence the raw features are separate in Figure~\ref{fig:encode}. In contrast, the adversarial training process successfully guides the spatial encoder to learn more uniformly distributed node embeddings on different graphs.


\begin{figure}[h]
    \centering
    \includegraphics[width=0.475\textwidth]{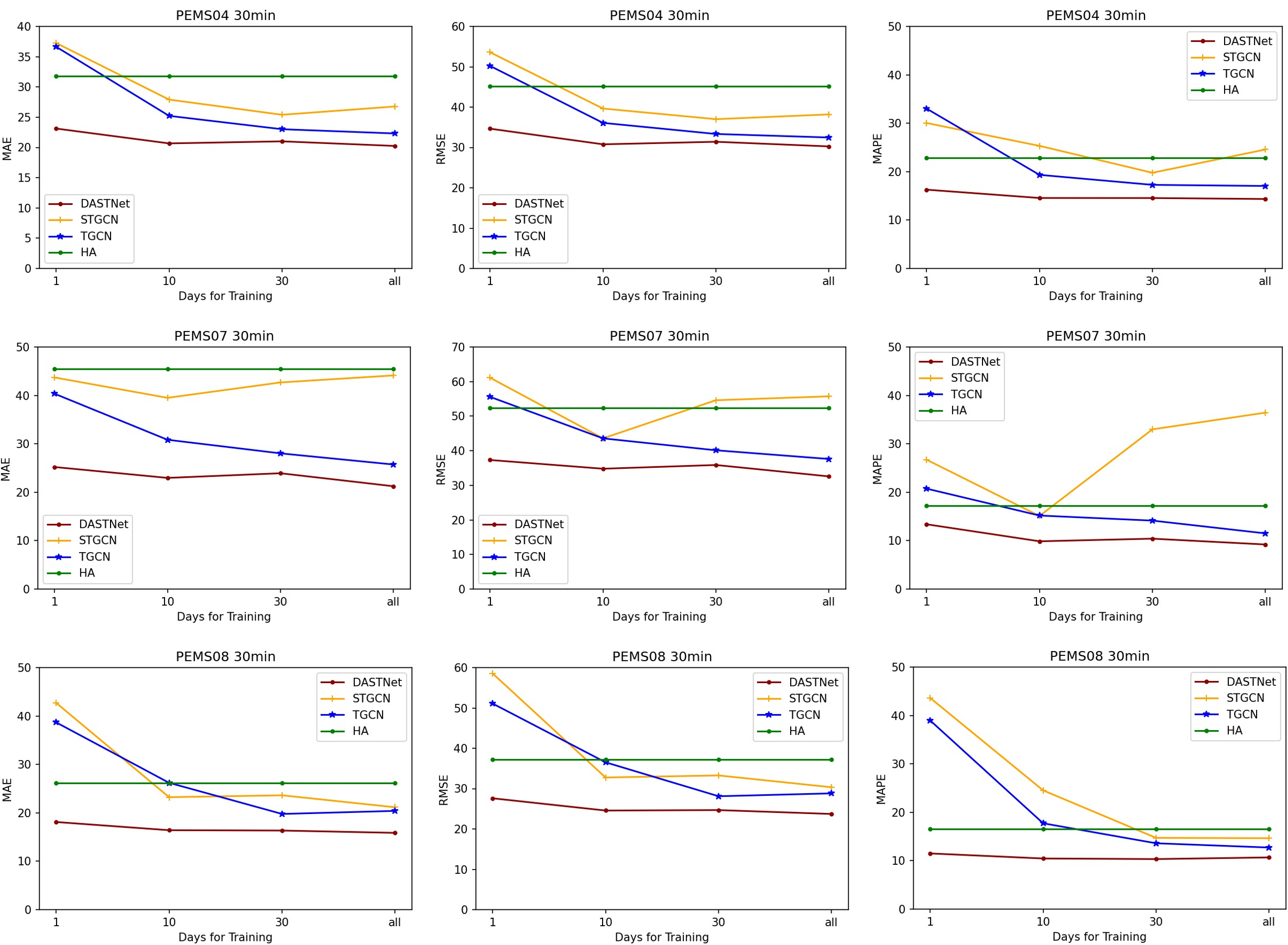}
    \caption{Sensitivity analysis, future 30-minute traffic flow forecasting results under different training set sizes.}
    \label{fig:sensitivity_MAE}
\end{figure}

\noindent\textbf{Sensitivity Analysis.}
To further demonstrate the robustness of \textsc{DastNet}, we conduct additional experiments with different sizes of training sets. We change the number of days of traffic flow data in the training set. To be more specific, we use four training sets with 1 day, 10 days, 30 days and all data, respectively. Then we compare \textsc{DastNet} with \textsc{Stgcn} and \textsc{Tgcn}. The performance of \textsc{Dcrnn} degrades drastically when the training set is small. To ensure the complete display in the Figure, we do not include it in the comparison and we do not include \textsc{Stgode} because of its instability.
We measure the performance of \textsc{DastNet} and the other two models on PEMS04, PEMS07, and PEMS08, by changing the ratio (measured in days) of the traffic flow data contained in the training set. 

Experimental results of the sensitivity analysis are provided in Figure \ref{fig:sensitivity_MAE}. In most cases, we can see that \textsc{Stgcn} and \textsc{Tgcn} underperform \textsc{Ha} when the training set is small. On the contrary, \textsc{DastNet} consistently outperforms other models in predicting different future time intervals of all datasets. Another observation is that the improvements over baseline methods are more significant for few-shot settings (small training sets). Specifically, the approximate gains on MAE decrease are 42.1\%/ 23.3\% /14.7\% /14.9\% on average for 1/10/30/all days for training compared with \textsc{Tgcn} and 46.7\%/35.7\% /30.7\% /34\% compared with \textsc{Stgcn}. 


\begin{figure}[h]
    \centering
    \includegraphics[width=0.45\textwidth]{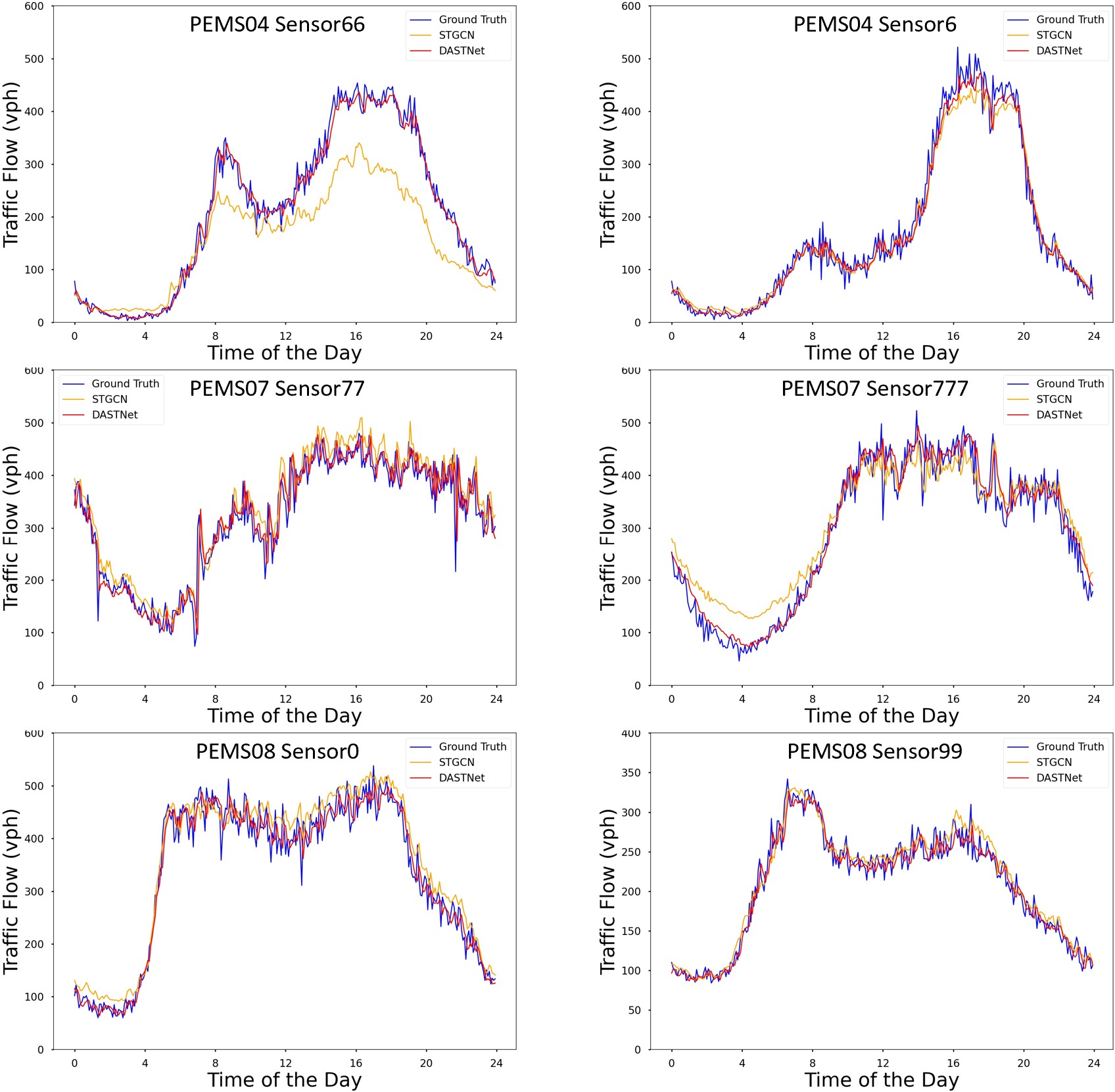}
    \caption{Visualization of the predicted flow.}
    \label{fig:casestudy}
\end{figure}

\noindent\textbf{Case Study.}
We randomly select six detectors and visualize the predicted traffic flow sequences of \textsc{DastNet} and \textsc{Stgcn} follow the setting in \cite{fang2021spatial}, and the visualizations are shown in Figure \ref{fig:casestudy}. Ground true traffic flow sequence is also plotted for comparison. One can see that the prediction generated by \textsc{DastNet} are much closer to the ground truth than that by \textsc{Stgcn}. \textsc{Stgcn} could accurately predict the peak traffic , which might be because \textsc{DastNet} learns the traffic trends from multiple datasets and ignores the small oscillations that only exist in a specific dataset.




\begin{figure*}[tb]
    \centering
    \includegraphics[width=0.94\textwidth]{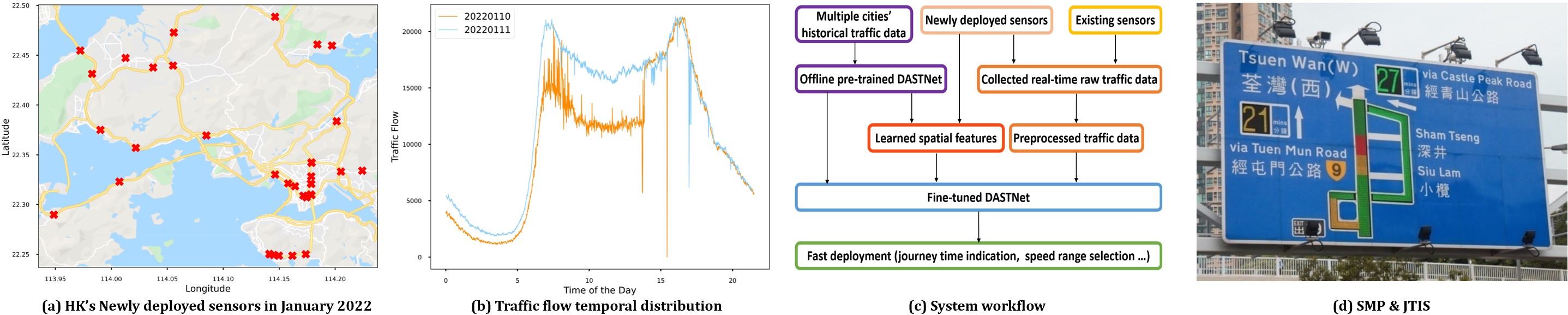}
    \caption{Traffic data and system workflow for the experimental deployment of \textsc{DastNet} in Hong Kong.}
    \label{fig:HKst}
\end{figure*}

\subsection{Experimental Deployment in Hong Kong}

By the end of 2022, we aim to deploy a traffic information provision system in Hong Kong using traffic detector data on strategic routes from the Transport Department \cite{transportdepartment}. 
The new system could supplement the existing Speed Map Panels (SMP) and Journey Time Indication System (JTIS) by employing more reliable models and real-time traffic data. 
For both systems, flow data is essential and collected from traffic detectors at selected locations for the automatic incident detection purpose, as the JTIS and SMP make use of the flow data to simulate the traffic propagation, especially after car crashes \cite{tam2011application}. Additionally, \textsc{DastNet} could be further extended for speed forecasting. 
As we discussed in Section~\ref{sec:intro}, the historical traffic data for the new detectors in Hong Kong are very limited. 
Figure \ref{fig:HKst} demonstrates: a) the spatial distribution of the newly deployed detectors in January 2022 and b) the corresponding traffic flow in Hong Kong. After the systematic process of the raw data as presented in c), traffic flow on the new detectors can be predicted and fed into the downstream applications once the detector is available.


We use the traffic flow data from three PEMS datasets for pre-training, and use Hong Kong's traffic flow data on January 10, 2022 to fine-tune our model. All Hong Kong's traffic flow data on January 11, 2022 are used as the testing set. We use 614 traffic detectors (a combinations of video detectors and automatic licence plate recognition detectors) to collect Hong Kong's traffic flow data for the deployment of our system, and the raw traffic flow data is aggregated to 5-minute intervals. We construct Hong Kong's road network $\mathcal{G}_{HK}$ based on distances between traffic detectors and define the adjacency matrix through connectivity.. Meanwhile, \textsc{Ha} and spatial-temporal baseline methods \textsc{Tgcn}, \textsc{Stgcn} and \textsc{Stgode} are adopted for comparisons. 
All experiments are repeated for 5 times, and the average results are shown in Table \ref{table:hkfull}. One can read from the table that, with the trained \textsc{DastNet} from other datasets, accurate traffic predictions can be delivered to the travelers immediately (after one day) when the detector data is available.

\renewcommand{\arraystretch}{0.95}
\begin{table}[ht]
\setlength{\tabcolsep}{1pt}
\caption{Performance comparison on the newly collected data in Hong Kong.}
\centering
\resizebox{0.43\textwidth}{!}{
\begin{tabular}{c||ccc|ccc|ccc}
\hline \hline
\multirow{2}{*}{HK} & \multicolumn{3}{c|}{15min} & \multicolumn{3}{c|}{30min} & \multicolumn{3}{c}{60min} \\ \cline{2-10}
                        & MAE    & RMSE    & MAPE   & MAE    & RMSE    & MAPE   & MAE    & RMSE    & MAPE   \\ \hline\hline
\textsc{Ha}                  & 15.79    & 23.95        & 16.96\%       & 17.84    & 27.00        & 18.70\%       & 21.66       & 33.41        & 22.42\%     \\ \hline
\textsc{Tgcn}                  & 22.39    & 30.50        & 27.54\%       & 22.39       & 30.48        & 26.76\%       & 25.95       & 35.61        & 27.98\%     \\ \hline
\textsc{Stgcn}                  & 39.86    & 55.79        & 46.80\%       & 39.34       & 55.34        & 45.62\%       & 42.52       & 58.95        & 52.94\%     \\ \hline
\textsc{Stgode}                  & 63.46    & 86.08        & 54.77\%       & 66.19       & 87.36        & 69.23\%       & 66.76       & 92.83        & 58.65\%     \\ \hline

\textsc{DastNet}             & \textbf{11.71}    & \textbf{17.69}        & \textbf{12.89}\%        & \textbf{13.87}       & \textbf{21.25}        & \textbf{14.91}\%       & \textbf{17.09}       & \textbf{26.47}        & \textbf{18.24}\%       \\ 
\hline \hline
\end{tabular}}
\label{table:hkfull}
\end{table}

\section{Conclusion}\label{sec:conclusion}

In this study, we formulated the transferable traffic forecasting problem and proposed an adversarial multi-domain adaptation framework named Domain Adversarial Spatial-Temporal Network (\textsc{DastNet}). 
This is the first attempt to apply adversarial domain adaptation to network-wide traffic forecasting tasks on the general graph-based networks to the best of our knowledge. 
Specifically, \textsc{DastNet} is pre-trained on multiple source datasets and then fine-tuned on the target dataset to improve the forecasting performance. 
The spatial encoder learns the uniform node embedding for all graphs, the domain classifier forces the node embedding domain-invariant, and the temporal forecaster generates the prediction. 
\textsc{DastNet} obtained significant and consistent improvements over baseline methods on benchmark datasets and will be deployed in Hong Kong to enable the smooth transition and early deployment of smart mobility applications. 

We will further explore the following aspects for future work: (1) Possible ways to evaluate, reduce and eliminate discrepancies of time-series-based graph signal sequences across different domains. (2) The effectiveness of the private encoder does not conform to domain adaptation theory \cite{ben2010theory}, and it is interesting to derive theoretical guarantees for the necessity of the private encoder on target domains.
In the experimental deployment, we observe that the performance of existing traffic forecasting methods degrades drastically when the traffic flow rate is low. However, this situation is barely covered in the PEMS datasets, which could potentially make the current evaluation of traffic forecasting methods biased. (3) The developed framework can potentially be utilized to learn the node embedding for multi-tasks, such as forecasting air pollution, estimating population density, etc. It would be interesting to develop a model for a universal location embedding \cite{bommasani2021opportunities}, which is beneficial for different types of location-related learning tasks \cite{wu2020learning, chen2021robust}.


\section*{Acknowledgments}
This study was supported by the Research Impact Fund for ``Reliability-based Intelligent Transportation Systems in Urban Road Network with Uncertainty'' and the Early Career Scheme from the Research Grants Council of the Hong Kong Special Administrative Region, China (Project No. PolyU R5029-18 and PolyU/25209221), as well as a grant from the Research Institute for Sustainable Urban Development (RISUD) at the Hong Kong Polytechnic University (Project No. P0038288). The authors thank the Transport Department of the Government of the Hong Kong Special Administrative Region for providing the relevant traffic data.



\clearpage

\bibliographystyle{ACM-Reference-Format}
\balance
\bibliography{main}

\end{document}